\documentclass[11pt,a4paper]{article}
\usepackage[hyperref]{acl2020}
\hypersetup{
    linkcolor=magenta,
    filecolor=magenta,
}

\usepackage{times}
\usepackage{latexsym}

\usepackage{microtype}

\aclfinalcopy

\usepackage{color,xcolor}
\usepackage{xspace}
\usepackage{graphicx}
\usepackage{adjustbox}
\usepackage{array}
\usepackage{booktabs}
\usepackage{colortbl}
\usepackage{float,wrapfig}
\usepackage{hhline}
\usepackage{multirow}
\usepackage{subcaption} 
\usepackage[toc,page]{appendix}
\usepackage{amsmath,amsfonts,amsthm,amssymb}
\usepackage{pifont}
\usepackage{multicol}
\definecolor{citecolor}{RGB}{34,139,34}
\definecolor{mydarkblue}{rgb}{0,0.08,1}
\definecolor{mydarkgreen}{rgb}{0.02,0.6,0.02}
\definecolor{mydarkred}{rgb}{0.8,0.02,0.02}
\definecolor{mydarkorange}{rgb}{0.40,0.2,0.02}
\definecolor{mypurple}{RGB}{111,0,255}
\definecolor{myred}{rgb}{1.0,0.0,0.0}
\definecolor{mygold}{rgb}{0.75,0.6,0.12}
\definecolor{myblue}{rgb}{0,0.2,0.8}
\definecolor{mydarkgray}{rgb}{0.,0.2,0.2}

\newcommand{\figref}[1]{Figure~\ref{#1}}
\newcommand{\tabref}[1]{Table~\ref{#1}}

\newcommand{\myparagraph}[1]{\vspace{-3pt}\paragraph{#1}}

\newcommand{\name}{HAT\xspace}
\newcommand{\x}{$\times$\xspace}

\newcommand{\supertrans}{SuperTransformer\xspace}
\newcommand{\supertranss}{SuperTransformers\xspace}
\newcommand{\subtrans}{SubTransformer\xspace}
\newcommand{\subtranss}{SubTransformers\xspace}

\newcommand{\multadd}{FLOPs\xspace}
\newcommand{\bleu}{BLEU\xspace}
\newcommand{\sacre}{SacreBLEU\xspace}

\newcommand{\cmark}{\ding{51}}%
\newcommand{\xmark}{\ding{55}}%

\newcommand{\wmtendebestmodelsizebase}{3.7}
\newcommand{\wmtenfrbestspeedupbase}{3}
\newcommand{\aedalatoverhead}{0.4}

\newcommand{\mycaption}[1]{\vspace{-16pt}\caption{#1}\vspace{-10pt}}
\newcommand{\tabcaption}[1]{\caption{#1}}

\newcommand{\wmtenfrbestmodelsizeet}{3.6}
\newcommand{\wmtenfrbestspeedupet}{2.7}

\newcommand{\lesssearchcostet}{12,041}

\newcommand{\levenspeedup}{1.9}

\title{HAT: \underline{H}ardware-\underline{A}ware \underline{T}ransformers for \\
Efficient Natural Language Processing}

\author{
Hanrui Wang$^1$, Zhanghao Wu$^1$, Zhijian Liu$^1$, Han Cai$^1$, Ligeng Zhu$^1$, \\ 
\textbf{Chuang Gan$^2$, Song Han$^1$} \\
$^1$Massachusetts Institute of Technology, \quad $^2$MIT-IBM Watson AI Lab \\
\small{\texttt{\{hanrui,zhwu,zhijian,hancai,ligeng,chuangg,songhan\}@mit.edu}}
}

\date{}

\begin{document}
\maketitle
\begin{abstract}

Transformers are ubiquitous in Natural Language Processing (NLP) tasks, but they are difficult to be deployed on hardware due to the intensive computation. To enable low-latency inference on resource-constrained hardware platforms, we propose to design Hardware-Aware Transformers (HAT) with neural architecture search.
We first construct a large design space with \emph{arbitrary encoder-decoder attention} and \emph{heterogeneous layers}.
Then we train a \textit{\supertrans} that covers all candidates in the design space, and efficiently produces many \emph{\subtranss} with weight sharing.
Finally, we perform an evolutionary search with a hardware latency constraint to find a specialized \emph{\subtrans} dedicated to run fast on the target hardware.
Extensive experiments on four machine translation tasks demonstrate that \name can discover efficient models for different hardware (CPU, GPU, IoT device).
When running WMT'14 translation task on Raspberry Pi-4, \name can achieve \textbf{\wmtenfrbestspeedupbase\x} speedup, \textbf{\wmtendebestmodelsizebase}\x smaller size over baseline Transformer; \textbf{\wmtenfrbestspeedupet}\x speedup, \textbf{\wmtenfrbestmodelsizeet}\x smaller size over Evolved Transformer with \textbf{\lesssearchcostet}\x less search cost and no performance loss. HAT is open-sourced at \href{https://github.com/mit-han-lab/hardware-aware-transformers.git}{\color{magenta}{https://github.com/mit-han-lab/hardware-aware-transformers.git}}.

\end{abstract}

\section{Introduction}

Transformer~\cite{Vaswani:2017attention} has been widely used in natural language processing tasks.
By stacking multiple identical encoder/decoder layers with attention modules, it provides a significant performance improvement over previous convolutional or recurrent neural network models~\cite{kim2014convolutional}.

\begin{figure}[t]
    \centering
    \includegraphics[width=\columnwidth]{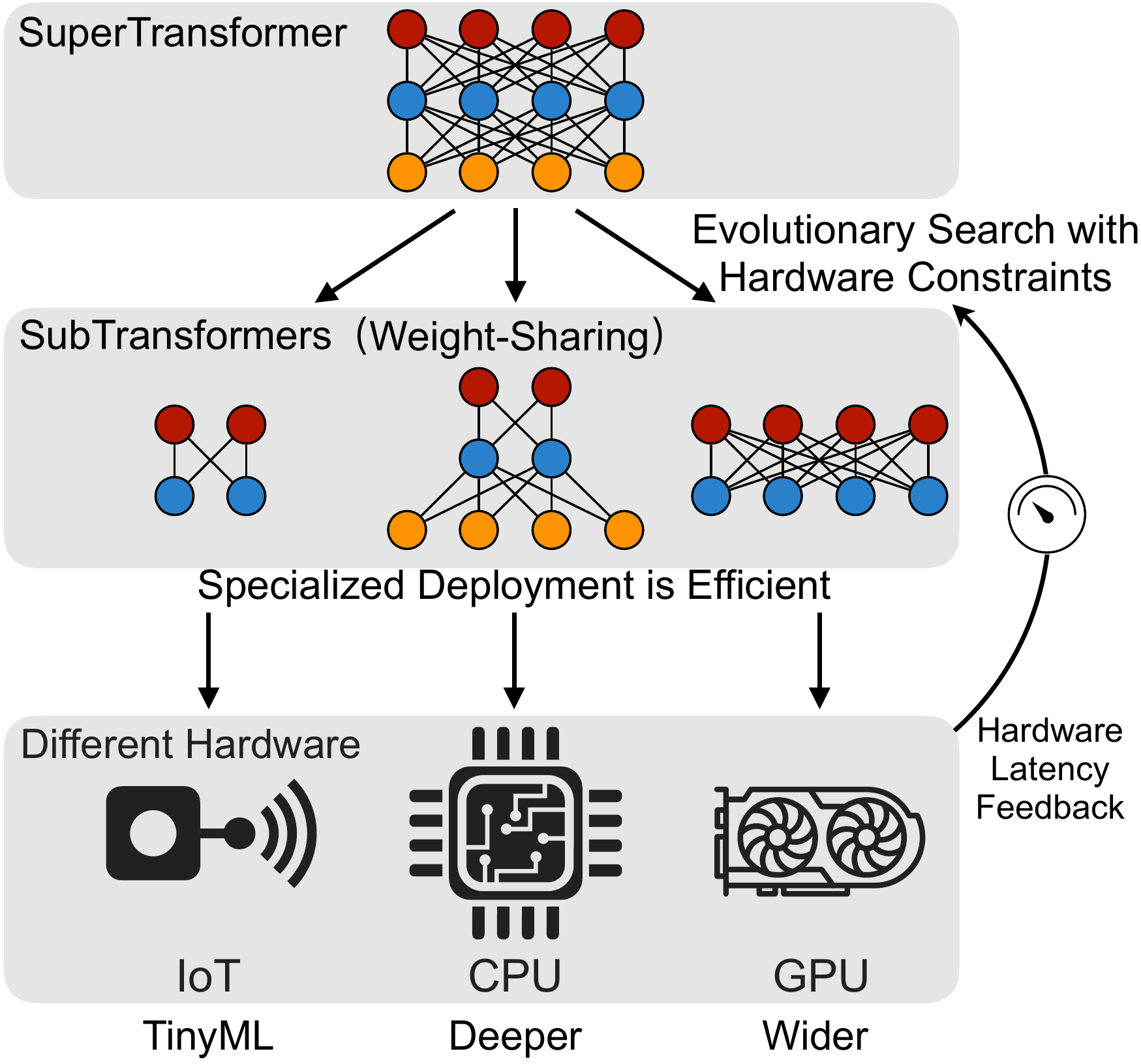}
    \vspace{-5pt}
    \mycaption{Framework for searching Hardware-Aware Transformers. We first train a \supertrans that contains numerous sub-networks, then conduct an evolutionary search with hardware latency feedback to find one \textbf{specialized} \subtrans for each hardware.}
    \vspace{-5pt}
    \label{fig:teaser}
\end{figure}

\begin{figure*}[t]
    \centering
    \includegraphics[width=\textwidth]{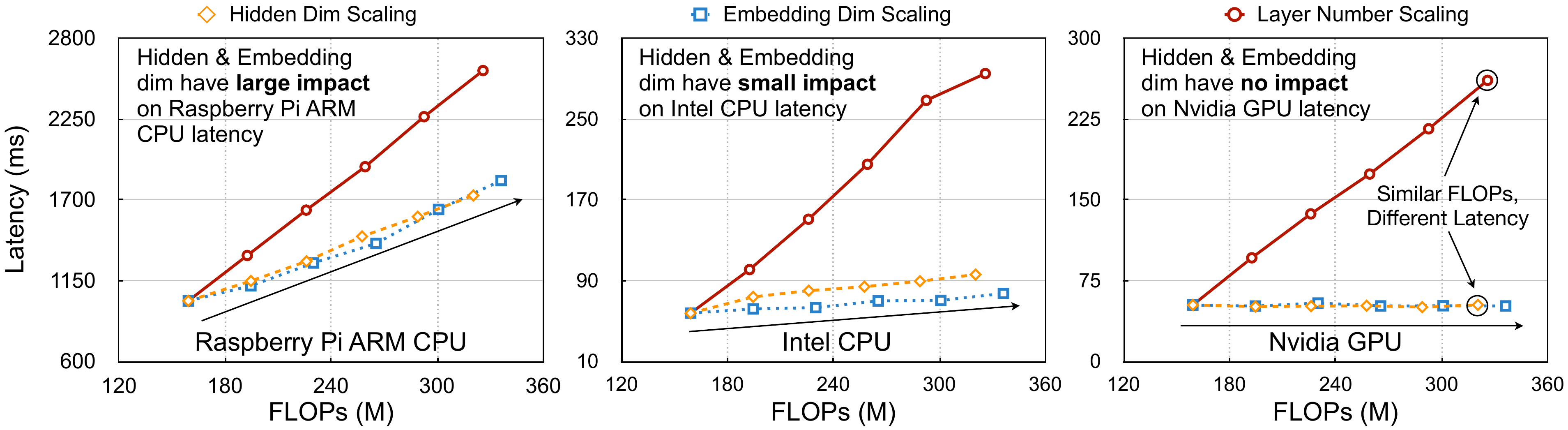}
    \vspace{-5pt}
    \mycaption{Latency of different Transformer models on different hardware. We find (1) FLOPs does not reflect the real measured latency; (2) Latency influencing factors of different hardware are contrasting. Thus we need to consider hardware latency feedback to design specialized models for different hardware.
    }
    \vspace{-5pt}
    \label{fig:latency_profile}
\end{figure*}

Nevertheless, it is challenging to deploy Transformers on mobile devices due to the high computation cost.
For instance, in order to translate a sentence with only 30 words, a Transformer-Big model needs to execute
13G \multadd and takes 20 seconds on a Raspberry Pi. Such long latency will hurt the user experience on edge devices. Thus we need hardware-efficient Transformers (Figure~\ref{fig:teaser}).

There are two common pitfalls when evaluating the efficiency of a Transformer.
\emph{(1)} \emph{FLOPs does not reflect the measured latency}. Although FLOPs is used as an metric for efficiency in prior arts~\cite{Howard:2017mobilenets, Anonymous:2020efficient}, it is not a good latency proxy. As in Figure~\ref{fig:latency_profile} (\emph{Right}), models with the \emph{same} FLOPs can result in very \emph{different} measured latencies;
\emph{(2)} \emph{different hardware prefers different Transformer architecture.} 
As in~\tabref{tab:diagonal}, the Transformer model optimized on one hardware is \emph{sub-optimal} for another because latency is influenced by different factors on different hardware platforms.
For example, the embedding dimension has significant impact on the Raspberry Pi latency but hardly influences the GPU latency (Figure~\ref{fig:latency_profile}).

\begin{table}[t]
    \renewcommand*{\arraystretch}{1.2}
    \small\centering
    \begin{tabular}{lc|c|c}
        \toprule
        \multicolumn{2}{r}{\small \underline{Measured On $\rightarrow$}} &  \multicolumn{1}{c}{GPU}     & ARM CPU \\
        {\small \underline{Specialized For $\downarrow$}} & \multicolumn{1}{c}{BLEU} &  \multicolumn{1}{c}{Latency} & \multicolumn{1}{c}{Latency} \\
        \midrule
        HAT (GPU) & 28.10 & \cellcolor{blue!20}\textbf{147 ms}  & 
        \cellcolor{myred!20}6491 ms \\
        HAT (ARM CPU) & 28.15 & \cellcolor{myred!20}184 ms & \cellcolor{blue!20}\textbf{6042 ms} \\
        \bottomrule
    \end{tabular}
    \vspace{-5pt}
    \tabcaption{\bleu score and measured inference latency of HAT on WMT'14 En-De task. The efficient model for GPU is not efficient for ARM CPU and vice versa. 
    }
    \vspace{-18pt}
    \label{tab:diagonal}
\end{table}

Inspired by the success of Neural Architecture Search (NAS)~\cite{pmlr-v80-bender18a, Guo:2019single, Pham:2018tl, cai2019once}, we propose to search for \textbf{H}ardware-\textbf{A}ware \textbf{T}ransformers (\name) by \emph{directly} involving the latency feedback into the design loop. In this way, we do not need FLOPs as the latency proxy and can search specialized models for various hardware. 

We first construct a large search space with \emph{arbitrary encoder-decoder attention} and \emph{heterogeneous Transformer layers}. 
Traditional Transformer has an information bottleneck between the encoder and decoder. Arbitrary encoder-decoder attention breaks the bottleneck, allowing all decoder layers to attend to multiple and different encoder layers instead of only the last one. Thus low-level information from the encoder can also be used by the decoder. Motivated by Figure \ref{fig:latency_profile}, we introduce heterogeneous Transformer layers to allow different layers to have different architecture adapting various hardware.

To perform a low-cost search in such a large design space, we first train a Transformer supernet -- \supertrans, which contains many \subtranss sharing the weights. We train all \subtranss simultaneously by optimizing the uniformly sampled \subtranss from the \supertrans.
The performance of a \subtrans with inherited weights from the \supertrans can provide a good relative performance approximation for different architectures trained from-scratch. Unlike conventional NAS, we only need to pay the \supertrans training cost for \textit{once} and can evaluate \textit{all} the models in the design space with it.
Finally, we conduct an evolutionary search to find the best \subtrans under the hardware latency constraint. Experiments show that \name can be naturally incorporated with model compression techniques such as quantization and knowledge distillation.

We evaluate \name with WMT'14 En-De, WMT'14 En-Fr, WMT'19 En-De, and IWSLT'14 De-En tasks on Raspberry Pi ARM CPU, Intel Xeon CPU, and Nvidia TITAN Xp GPU. Compared with previous work~\cite{Vaswani:2017attention,So:2019et,Gu:2019levenshtein,Anonymous:2020efficient}, \name achieves up to \textbf{\wmtenfrbestspeedupbase\x} speedup, \textbf{\wmtendebestmodelsizebase}\x smaller size over Transformer-Big without loss of accuracy. With \textbf{12,041$\times$} less search cost, \name outperforms the Evolved Transformer with \textbf{\wmtenfrbestspeedupet}\x speedup and \textbf{\wmtenfrbestmodelsizeet}\x smaller size. It also achieves up to \textbf{\levenspeedup}\x speedup over Levenshtein and Lite Transformers with no \bleu score loss. With 4-bit quantization, \name can further reach \textbf{25\x} model size reduction.

\begin{figure*}[t]
    \centering
    \includegraphics[width=\textwidth]{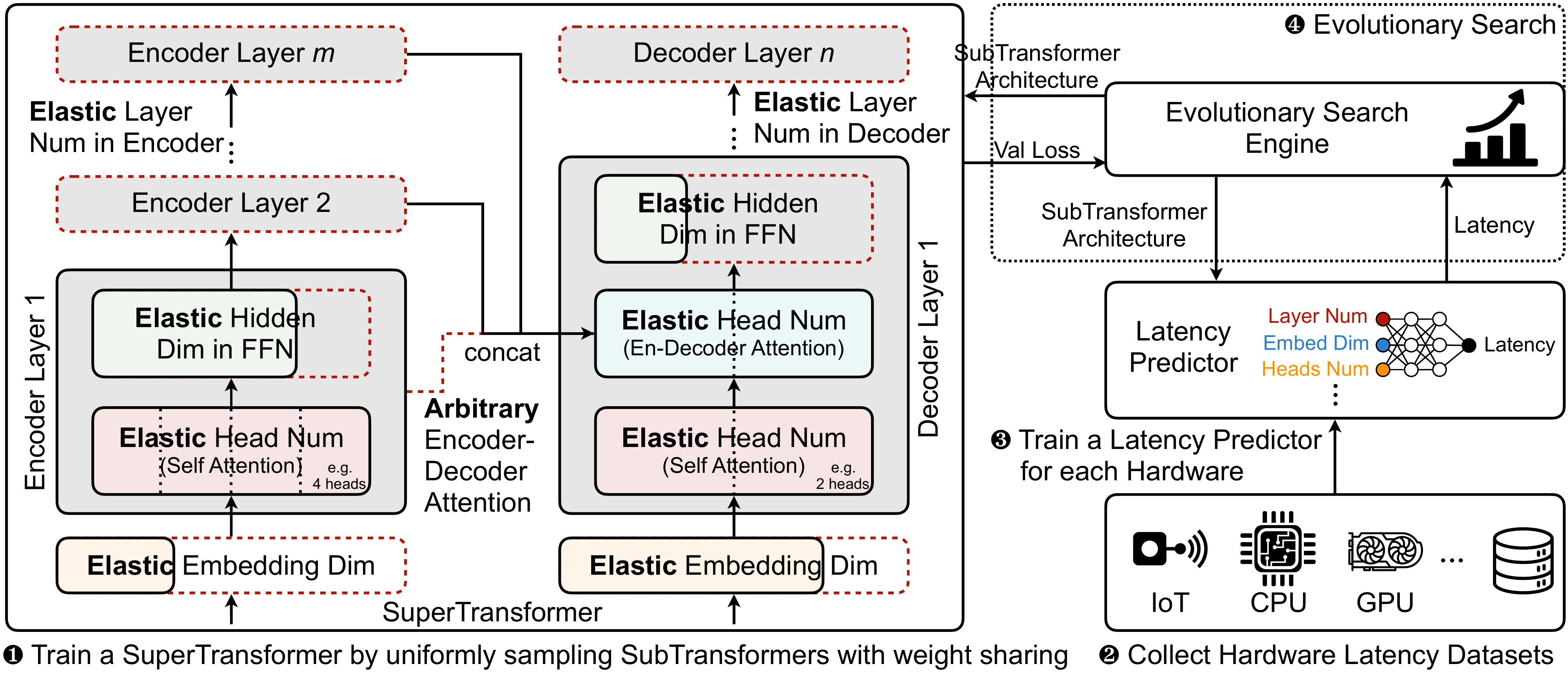}
    \vspace{-4pt}
    \mycaption{\name Overview. A large design space is constructed with Arbitrary Encoder-Decoder Attention and Heterogeneous Layers. %
    (1) Train a weight-shared \supertrans by iteratively optimizing randomly sampled \subtranss. It can provide a performance proxy for \subtranss. %
    (2) Collect \emph{(\subtrans architecture, latency)} data pairs on the target hardware. %
    (3) Train a latency predictor for each hardware to provide fast and accurate latency feedback. %
    (4) Perform an evolutionary search with hardware latency constraint to find the model with the \emph{lowest validation loss}. %
    (5) Finally, the searched model is trained \emph{from scratch} to get the final performance.}
    \vspace{-4pt}
    \label{fig:overview}
\end{figure*}

\name has three contributions: (1) \textbf{Hardware-Aware and Specialization.} To our best knowledge, we are the first to directly involve the hardware feedback in the model design, to reduce NLP model latency for target hardware, instead of relying on proxy signals (FLOPs). For different hardware platforms, specialized models for low-latency inference are explored. (2) \textbf{Low-cost Neural Architecture Search with a Large Design Space.} We propose \emph{arbitrary encoder-decoder attention} to break the information bottleneck; and \emph{heterogeneous layer} to let different layers alter its capacity. A weight-shared \supertrans is trained to search for efficient models at a low cost. (3) \textbf{Design Insights.} Based on the search results, we reveal some design insights: Attending to multiple encoder layers is beneficial for the decoder; GPU prefers shallow and wide models while ARM CPU prefers deep and thin ones.

\section{Proposed Approaches}

An overview of the \name framework is shown in~\figref{fig:overview}. We firstly train a \supertrans with a large design space. Then, for a given hardware platform, we collect a dataset of \emph{(\subtrans architecture, measured latency)} pairs for different models, and train a latency predictor. Finally, we conduct an evolutionary search with a latency constraint to find an efficient model specialized for the target hardware.

\subsection{Design Space}

We construct a large design space by breaking two conventions in the Transformer design: (1) All decoder layers only attend to the last encoder layer; (2) All the layers are identical.

\begin{figure}[t]
    \centering
    \includegraphics[width=\columnwidth]{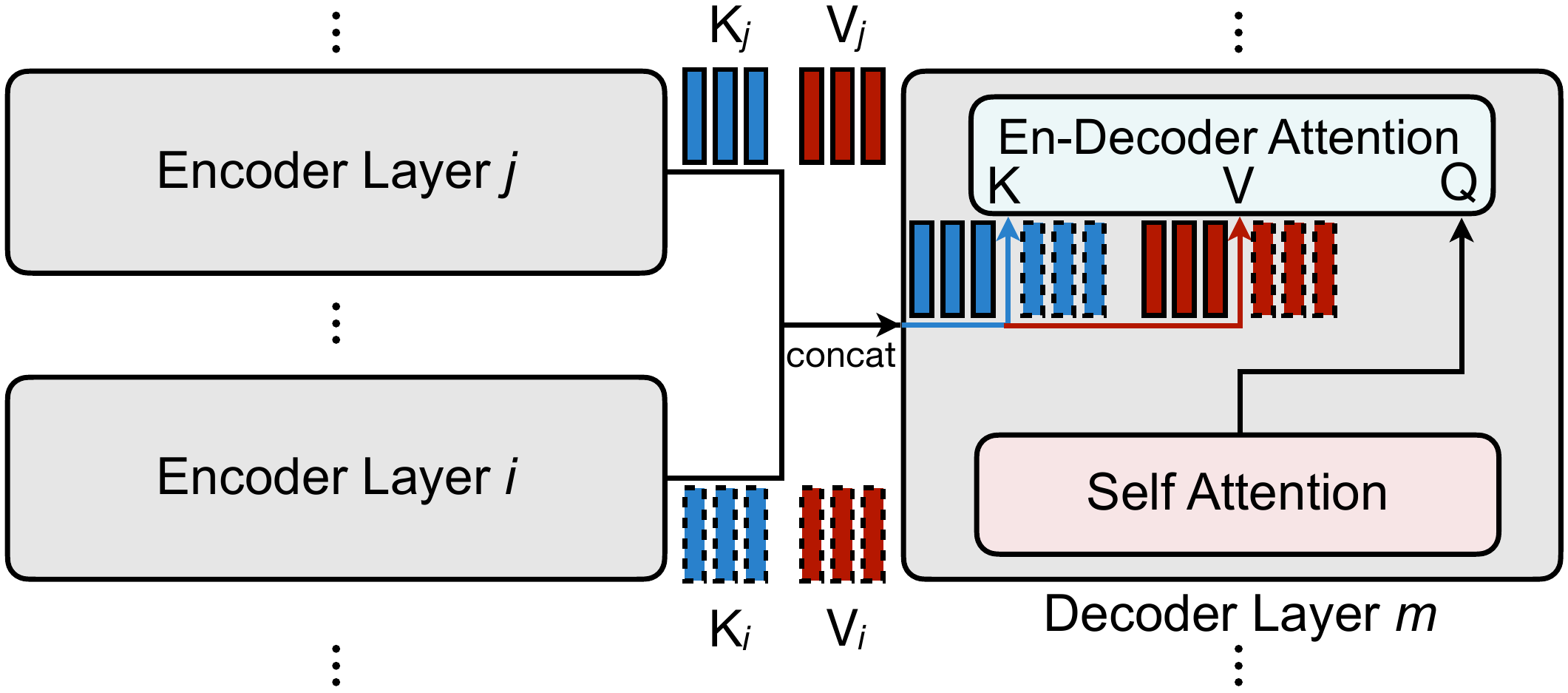}
    \vspace{-6pt}
    \mycaption{Arbitrary Encoder-Decoder Attention. Each encoder-decoder attention in one decoder layer can attend to the outputs from multiple encoder layers, fully leveraging the features extracted by the encoder.
    }
    \label{fig:ende_attn}
\end{figure}

\myparagraph{Arbitrary Encoder-Decoder Attention.}
Different encoder layers extract features on different abstraction levels. Conventionally, all the decoder layers only attend to the last encoder layer. It forms an \emph{information bottleneck} that forces all the decoder layers to learn solely from the high abstraction level and ignore the low-level information. To break the bottleneck, we propose Arbitrary Encoder-Decoder Attention to learn the most suitable connections between the encoder and the decoder. Each decoder layer can choose \emph{multiple} encoder layers to attend. The \emph{key} and \emph{value} vectors from encoder layers are concatenated in the \emph{sentence length dimension} (\figref{fig:ende_attn}) and fed to the encoder-decoder cross attention module. The mechanism is efficient because it introduces no additional parameters. The latency overhead is also negligible. For example, with each decoder layer attending to two encoder layers, the latency of Transformer-Base on Nvidia TITAN Xp GPU barely increases by  \aedalatoverhead\%. It improves the model capacity by allowing attention to different abstraction levels.

\myparagraph{Heterogeneous Transformer Layers.}
Previous Transformers repeat one architecture for all layers. In \name, instead, different layers are \emph{heterogeneous}, with different numbers of heads, hidden dim, and embedding dim. In attention layers, different heads are used to capture various dependencies. However, ~\citet{Voita:2019analyzing} shows that many heads are redundant. We thereby make attention head number \emph{elastic} so that each attention module can decide its necessary number of heads. 

In the FFN layer, the input features are cast to a higher dimension (hidden dim), followed by an activation layer. Traditionally, the hidden dim is set as 2\x or 4\x of the embedding dim, but this is sub-optimal since different layers need different capacities depending on the feature extraction difficulty. We hence make the hidden dim \emph{elastic}.

Moreover, we also support \emph{elastic} embedding dim of encoder and decoder, but it is consistent inside encoder/decoder. The number of encoder \& decoder layers are also \emph{elastic} to learn the proper level of feature encoding and decoding. Other design choices such as the length of $Q, K, V$ vectors in attention modules can be naturally incorporated in our framework, which we leave for future work.

\begin{figure}[t]
    \centering
    \includegraphics[width=0.95\columnwidth]{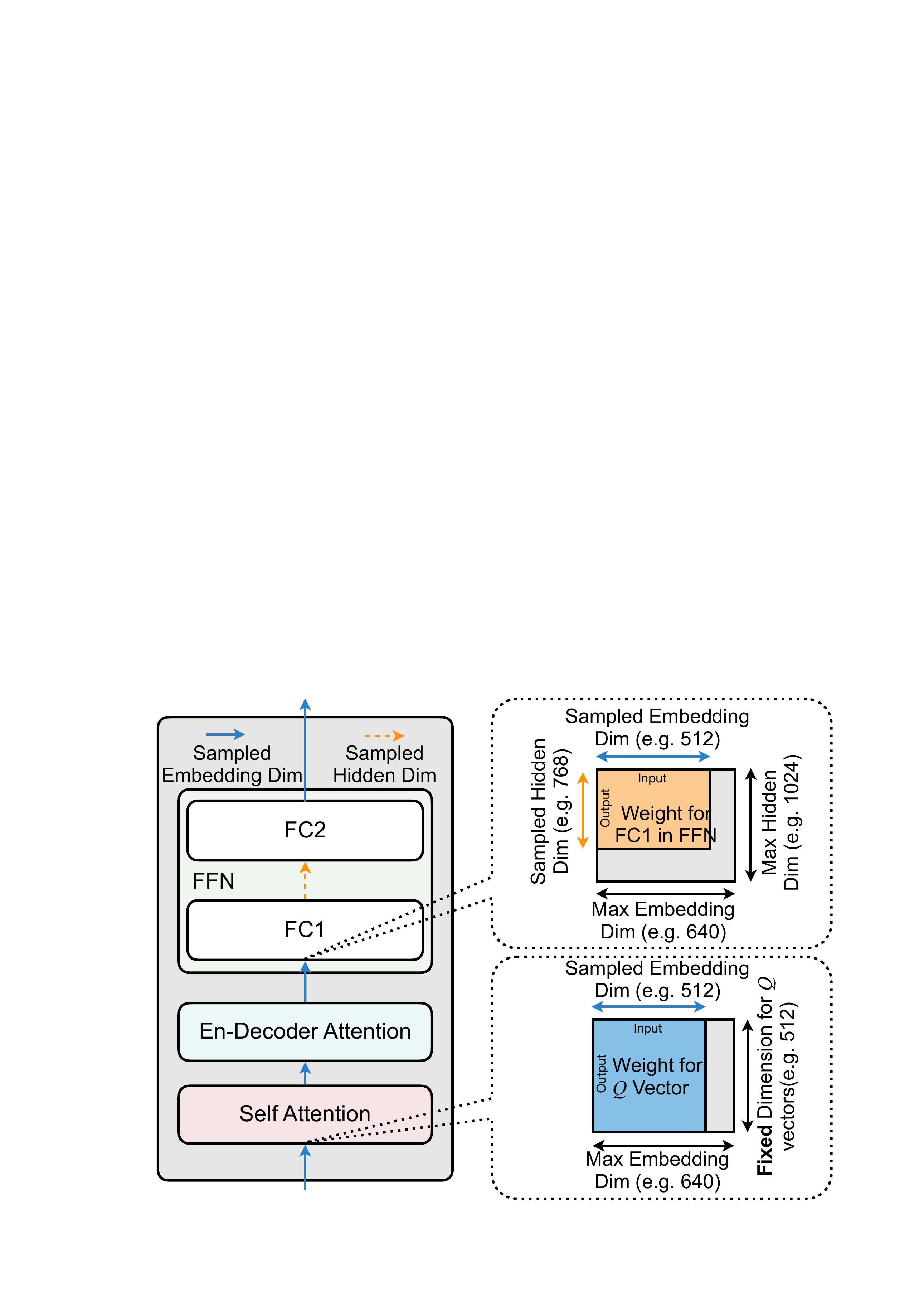}
    \vspace{-5pt}
    \caption{Weight Sharing of the \supertrans. All \subtranss share the front portion of word embeddings, and weights in the fully-connected layers.}
    \vspace{-15pt}
    \label{fig:weight_sharing}
\end{figure}

\subsection{\supertrans}

It is critical to have a large design space in order to find high-performance models.
However, training all the models and comparing their \bleu scores is infeasible. We thus propose \supertrans, a supernet for \emph{performance approximation}, which can judge the performance of a model without fully training it. The \supertrans is the largest model in the search space with \emph{weight sharing} \cite{Pham:2018tl,liu2018darts,cai2019once}. Every model in the search space (a \subtrans) is a part of the \supertrans. All \subtranss share the weights of their common parts. For elastic embedding dim, all \subtranss share the front portion of the longest word embedding and corresponding FC layer weights. As in Figure~\ref{fig:weight_sharing}, for elastic FFN hidden dim, the front part of the FC weights is shared. For elastic head number in attention modules, the whole $Q, K, V$ vectors (the lengths are fixed in our design space) are shared by dividing into $head\_number$ parts. 
Elastic layer numbers let all \subtranss share the first several layers. 

In the \supertrans training, all possible \subtranss are \emph{uniformly sampled}, and the corresponding weights are updated. In practice, the \supertrans only needs to be trained for the same steps as a baseline Transformer model, which is fast and low-cost. After training, we can get the performance proxy of sampled models in the design space by evaluating the corresponding \subtranss on the validation set without training. 

\begin{figure}[t]
    \centering
    \includegraphics[width=0.7\columnwidth]{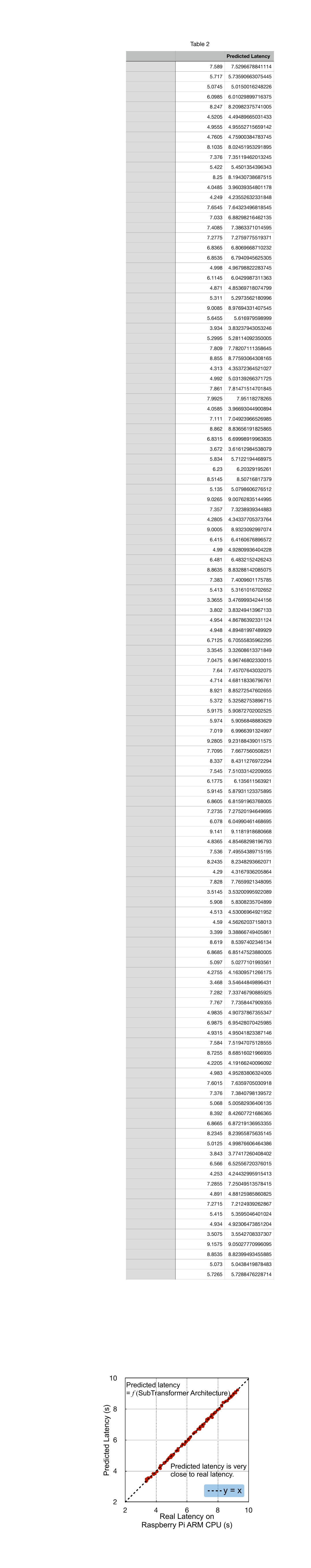}
    \vspace{6pt}
    \mycaption{The latency predictor is very accurate, with an average prediction error (RMSE) of 0.1s.}
    \label{fig:predicted_latency}
    \vspace{-5pt}
\end{figure}

\subsection{Evolutionary Search for \subtrans}
Given a latency requirement, we perform an evolutionary search to find a satisfactory \subtrans.
There are two ways to evaluate the hardware latency of a \subtrans: (1) Online measurement in which we measure the models during the search process. (2) Offline, where we train a \emph{latency predictor} to provide the latency. We apply the offline method here because it is \emph{fast and accurate}. For the online method, a single sampled \subtrans requires hundreds of inferences to get an accurate latency, which lasts for minutes and slows down the searching. For the offline method, we encode the architecture of a \subtrans into a feature vector, and predict its latency instantly with a multi-layer perceptron (MLP). Trained with thousands of real latency data points, the predictor yields high accuracy (Figure~\ref{fig:predicted_latency}). Note that the predicted latency is only used in the search process, and we report \emph{real measured latency} in the experiment section. Compared with deducing a closed-form latency model for each hardware, the latency predictor method is more general and faster.

We use an evolutionary algorithm to conduct the search process. As in Figure~\ref{fig:overview}, the search engine queries the latency predictor for \subtrans latency, and validates the loss on the validation set. The engine only adds \subtranss with latency \emph{smaller than} the hardware constraint to the population. We then train the searched models \emph{from scratch} to obtain the final performance.

\begin{figure*}[t]
    \centering
    \includegraphics[width=\textwidth]{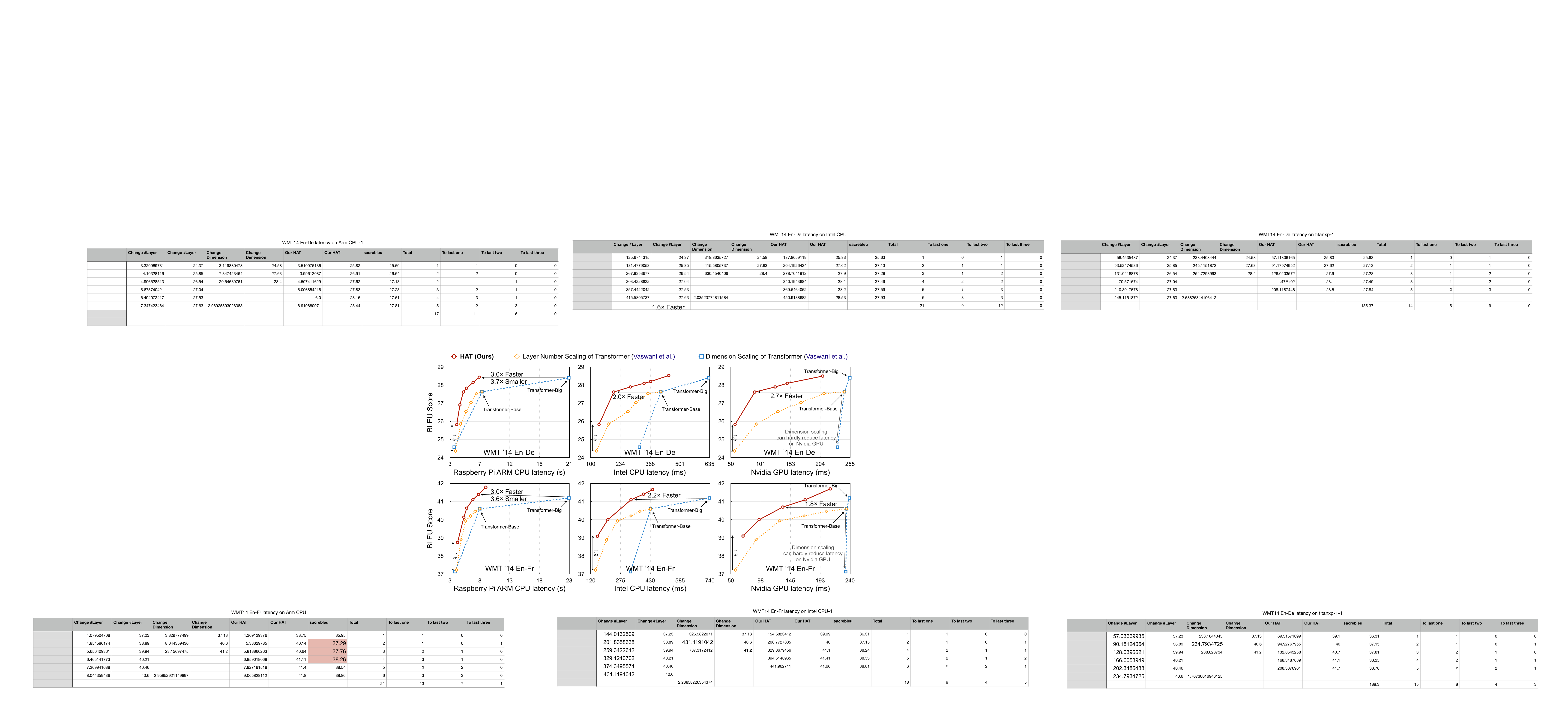}
    \mycaption{Inference latency and BLEU trade-offs of WMT'14 En-De and En-Fr on three hardware platforms. \name consistently outperforms the baseline Transformers and achieves up to \wmtenfrbestspeedupbase\x faster inference speed and \wmtendebestmodelsizebase\x smaller size over Transformer-Big. Specific latency, \bleu and \sacre~\cite{post2018call} are in Appendix Table~\ref{tab:appendix}.
    }
    \vspace{-8pt}
    \label{fig:tradeoff_curve}
\end{figure*}

\begin{figure}[t]
    \centering
    \includegraphics[width=\columnwidth]{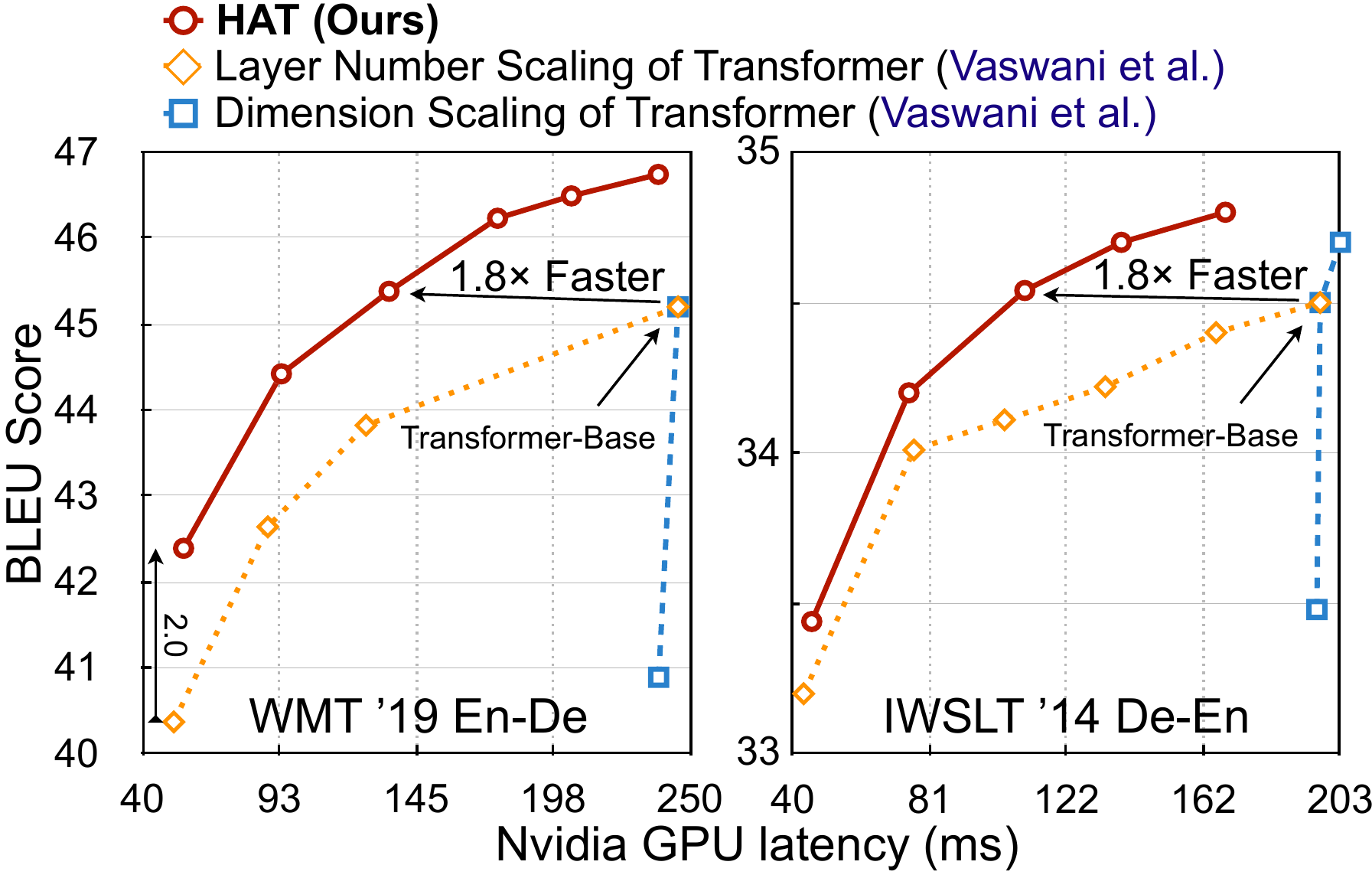}
    \vspace{-10pt}
    \caption{Inference latency and \bleu trade-offs of WMT'19 and IWSLT'14 tasks on Nvidia GPU.}
    \vspace{-16pt}
    \label{fig:19ende}
\end{figure}

\section{Experiments}

\subsection{Datasets}
We conduct experiments on four machine translation tasks: WMT'14 En-De, WMT'14 En-Fr, WMT'19 En-De, and IWSLT'14 De-En, consisting of 4.5M, 36.3M, 43.0M, and 160K pairs of training sentences, respectively. For WMT'14 En-De, we apply 32K source-target BPE vocabulary, train on WMT'16, validate on newstest2013 and test on newstest2014, replicating~\citet{Wu:2019payless}; For WMT'14 En-Fr, we use 40K source-target BPE vocabulary, validate on newstest2012\&2013, and test on newstest2014, replicating~\citet{Gehring:2017conv}. WMT'19 En-De adopts 49.6K source-target BPE vocabulary, validates on newstest2017, and tests on newstest2018, the same as~\citet{junczys2019microsoft}. We use 10K joint BPE vocabulary in lower case for IWSLT'14 De-En~\cite{Grave:2017ef}.

\subsection{Experiment Setups}
\label{sec:details}

\myparagraph{Baselines.}
Our baseline models are Transformer~\cite{Vaswani:2017attention}, Levenshtein Transformer~\cite{Gu:2019levenshtein}, both with the~\citet{Ott:2019fairseq} implementation, Evolved Transformer~\cite{So:2019et} and Lite Transformer~\cite{Anonymous:2020efficient}.

\myparagraph{Evaluation Metrics.}
For evaluation, we use beam four and length penalty 0.6 for WMT, and beam five for IWSLT~\cite{Vaswani:2017attention}. All BLEUs are calculated with case-sensitive tokenization\footnote{\href{https://github.com/moses-smt/mosesdecoder}{https://github.com/moses-smt/mosesdecoder}}, 
but we also apply the compound splitting BLEU\footnote{\href{https://github.com/tensorflow/tensor2tensor}{https://github.com/tensorflow/tensor2tensor}} for WMT, the same as~\citet{Vaswani:2017attention}. We test the model with the lowest validation set loss for WMT and the last ten checkpoints averaged for IWSLT.

We test the latency of the models by measuring translation from a source sentence to a target sentence with the same length. The length is the average output length on the test set -- 30 for WMT and 23 for IWSLT. For each model, we measure the latency for 300 times, remove the fastest and slowest 10\% and then take the average of the rest 80\%. We conduct experiments on three representative hardware platforms: Raspberry Pi-4 with an ARM Cortex-A72 CPU, Intel Xeon E5-2640 CPU, and Nvidia TITAN Xp GPU.

\begin{table*}[t]
\renewcommand*{\arraystretch}{1.15}
\setlength{\tabcolsep}{3pt}
\small\centering
\begin{tabular}{clccccccccc}
    \toprule
    & & \multirow{2}{*}{\shortstack[c]{Hardware-\\Aware}} & \multirow{2}{*}{\shortstack[c]{Hetero.\\Layers}} & \multirow{2}{*}{Latency}& \multirow{2}{*}{\#Params} & \multirow{2}{*}{\shortstack[c]{FLOPs\\(G)}}  & \multirow{2}{*}{BLEU} & \multirow{2}{*}{\shortstack[c]{GPU\\Hours}} & \multirow{2}{*}{\shortstack[c]{CO\textsubscript{2}e\\(lbs)}} & \multirow{2}{*}{\shortstack[c]{Cloud\\Comp. Cost}} \\
    &&&&&&&&&&\\
    
    \midrule
     \multirow{2}{*}{\shortstack[c]{IWSLT'14\\De-En}} & Transformer & \xmark & \xmark & 3.3s & 32M & 1.5  & 34.5 & 2 & 5 & \$12 - \$40  \\
     & \textbf{\name (Ours)} &\cmark &\cmark & \textbf{2.1s} & \textbf{23M}  & \textbf{1.1} & \textbf{34.5} & 4 & 9 & \$24 - \$80  \\
     
     \midrule

    \multirow{4}{*}{\shortstack[c]{WMT'14\\En-Fr}} & Transformer & \xmark & \xmark & 23.2s & 176M & 10.6 & 41.2 & 240 & 68 & \$178 - \$595 \\
    & Evolved Trans.   &\xmark & \xmark & 20.9s & 175M & 10.8 & 41.3 & 2,192,000 & 626,000 & \$1.6M - \$5.5M  \\
    & \textbf{\name (Ours)} &\cmark &\cmark & \textbf{7.8s} & \textbf{48M}  & \textbf{3.4} & \textbf{41.4} & 216 & 61 & \$159 - \$534  \\
    & \textbf{\name (Ours)} &\cmark &\cmark & 9.1s & 57M & 3.9 & \textbf{41.8} & 224 & 64 & \$166 - \$555  \\
    \midrule
        \multirow{4}{*}{\shortstack[c]{WMT'14\\En-De}} & Transformer &\xmark & \xmark & 20.5s & 176M  & 10.6 & 28.4 & 184 & 52 & \$136 - \$456  \\
    & Evolved Trans. &\xmark & \xmark & 7.6s & 47M & 2.9 & 28.2 & 2,192,000 & 626,000 & \$1.6M - \$5.5M  \\
    & \textbf{\name (Ours)} &\cmark &\cmark & \textbf{6.0s} & \textbf{44M} & \textbf{2.7} & 28.2 & 184 & 52 & \$136 - \$456  \\
    & \textbf{\name (Ours)} &\cmark &\cmark & 6.9s & 48M & 3.0   & \textbf{28.4} & 200 & 57 & \$147 - \$495 \\
    
    \bottomrule
\end{tabular}
\tabcaption{Comparisons of latency, model size, FLOPs, \bleu and training cost in terms of CO\textsubscript{2} emissions (lbs) and cloud computing cost (USD) for Transformer, the Evolved Transformer and \name. The training cost estimation is adapted from~\citet{Strubell:2019uv}. The training time is for one Nvidia V100 GPU, and the latency is measured on the Raspberry Pi ARM CPU. The cloud computing cost is based on AWS. 
}
\label{tab:evo_trans}
\end{table*}

\subsection{Implementation Details}
\myparagraph{\supertrans Setups.}
The \supertrans for WMT has the following design space: [512, 640] for embedding dim, [1024, 2048, 3072] for hidden dim, [4, 8] for the head number in all attention modules, [1, 2, 3, 4, 5, 6] for decoder layer number. Due to decoder auto-regression, encoder only accounts for less than 5\% of the measured latency; thereby, we set the encoder layer number fixed as 6. For arbitrary encoder-decoder attention, each decoder can choose to attend to the last one, two, or three encoder layers. The \supertrans design space for IWSLT is the same as WMT except for [2048, 1024, 512] for hidden dim and [4, 2] for head number. We set the $Q, K, V$ vector dim fixed as 512. The design space contains around $10^{15}$ possible \subtranss and covers a wide range of model size and latency (largest = 6$\times$smallest). We train the \supertranss of WMT for 40K steps and 50K steps for IWSLT.

\myparagraph{Hardware-Aware Evolutionary Search Setups.}
The input of the latency predictor is a feature vector of \subtrans architecture with ten elements: layer number, embed dim, average hidden dim, average self-attention heads, of both encoder and decoder; plus average encoder-decoder attention heads, and the average number of encoder layers each decoder layer attends. 
A dataset of 2000 (\subtrans architecture, measured latency) samples for each hardware is collected, and split into train:valid:test=8:1:1. We normalize the features and latency, and train a three-layer MLP with 400 hidden dim and ReLU activation. We choose three-layer because it is more accurate than the one-layer model, and over three layers do not improve accuracy anymore. With the predictor, we conduct an evolutionary search for 30 iterations in the \supertrans, with population 125, parents population 25, mutation population 50 with 0.3 probability and crossover population 50.

\myparagraph{Training Settings.}
Our training settings are in line with~\citet{Wu:2019payless} and~\citet{Anonymous:2020efficient}. For WMT, we train for 40K steps with Adam optimizer and a cosine learning rate (LR) scheduler~\cite{Kingma:2015adam,Loshchilov:2016sgdr}, where the LR is linearly warmed up from $10^{-7}$ to $10^{-3}$, and then cosine annealed. For IWSLT, we train for 50K steps with inverse square root LR scheduler. The baseline Transformers are trained with the \emph {same settings} as the searched \subtranss for fair comparisons.

\section{Results}
\subsection{\name Performance Comparisons}
\label{sec:comp}
In~\figref{fig:tradeoff_curve},~\ref{fig:19ende} and Appendix Table~\ref{tab:appendix}, we compare~\name with Transformer baselines on four tasks. The embedding dims are 512 and 1024 for the Transformer-Base and Big, respectively. The hidden dims are $4\times$ and $2\times$ of the embedding dim for WMT and IWSLT. The IWSLT models are smaller to prevent overfitting~\cite{Wu:2019payless}. We obtain a series of baseline models with layer number scaling (yellow) and dimension scaling (blue). We set different latency constraints on three hardware to get a series of \name models. \name consistently outperforms baselines with a large gap under different latency constraints. On ARM CPU, \name is 3\x faster and 3.7\x smaller than Transformer-Big with the same \bleu. On Intel CPU, \name achieves over 2\x speedup. On Nvidia GPU, the blue dash line is nearly \emph{vertical}, indicating that dimension scaling can hardly reduce the latency. In this case, \name can still find models with low latency and high performance.

We further compare various aspects of \name with Transformer ~\cite{Vaswani:2017attention} and Evolved Transformer~\cite{So:2019et} in~\tabref{tab:evo_trans}. \name achieves up to 1.6\x, 3\x, and 3.4\x speedup with up to 1.4\x, 3.7\x, and 4\x smaller size than baselines. We report FLOPs for translating a 23-token sentence for IWSLT and 30 for WMT. We show the overall GPU hours for training the \supertrans and the searched \subtrans.
We also calculate the cloud computing costs with different modes: ``preemptable" is cheaper (\$0.74/h) than ``on-demand" (\$2.48/h)~\cite{Strubell:2019uv}. \name is highly affordable since the total GPU-hour is over 12000\x smaller than the Evolved Transformer, and is even smaller than Transformer-Big by virtue of the compact model size.

\begin{table}[t]
    \renewcommand*{\arraystretch}{1.2}
    \setlength{\tabcolsep}{2pt}
    \small\centering
    \begin{tabular}{lcc}
        \toprule
        & Latency   & BLEU \\
        \midrule
        Transformer~\cite{Vaswani:2017attention}  & 4.3s & 25.85 \\
        Levenshtein~\cite{Gu:2019levenshtein} & 6.5s & 25.20 \\
        Evolved Transformer~\cite{So:2019et} & 3.7s  & 25.40 \\
        Lite Transformer~\cite{Anonymous:2020efficient} & 3.4s & 25.79 \\
        \midrule
        \textbf{HAT (Ours)} & \textbf{3.4s}  & \textbf{25.92} \\
        \bottomrule
    \end{tabular}
    \tabcaption{Raspberry Pi ARM CPU latency and \bleu comparisons with different models on WMT'14 En-De. 
    \name has the lowest latency with the highest \bleu.
    }
    \vspace{-10pt}
    \label{tab:comparison}
\end{table}

\begin{figure}[t]
    \centering
    \includegraphics[width=\columnwidth]{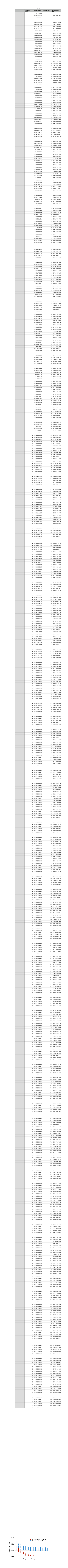}
    \mycaption{Evolutionary search can find better \subtranss in the \supertrans than random search.}
    \label{fig:evo_vs_random}
\end{figure}

In Table~\ref{tab:comparison}, we compare \name with other latest models. We scale down all models to have similar \bleu scores with Levenshtein for fair comparisons. We adopt the average iteration time of 2.88 for decoding~\cite{Gu:2019levenshtein}, without limiting the length of the output sentence (12 tokens after decoding). \name runs 1.3\x faster than Transformer with higher \bleu; 1.9\x faster than Levenshtein with 0.7 higher \bleu. Under similar latency, \name also outperforms Lite Transformer. These results demonstrate \name's effectiveness in lower latency scenarios. Our framework can also be adopted to speedup those models.

\subsection{Analysis}
\label{sec:analysis}
\myparagraph{Design Insights.}
For all \name WMT models in~Figure~\ref{fig:tradeoff_curve}, 10\% of all decoder layers attend to three encoder layers, 40\% attend to two encoder layers. That demonstrates
the necessity of arbitrary encoder-decoder attentions. 

In Appendix Figure~\ref{fig:supp}, we visualize the models specialized for different hardware mentioned in Table~\ref{tab:diagonal}. We find that the GPU model is \emph{wide but shallow}; the Raspberry Pi model is \emph{deep but thin}. The phenomenon echos with our latency profiling (Figure~\ref{fig:latency_profile}) as GPU latency is insensitive to embedding and hidden dim, but Raspberry Pi is highly sensitive. It guides manual designs: on GPU, we can reduce the layer number and increase dimension to reduce latency and keep high performance.

\myparagraph{Ablation Study.}
\begin{table}[t]
\renewcommand*{\arraystretch}{1.15}
\setlength{\tabcolsep}{3pt}
\small\centering
\begin{tabular}{llcccc}
    \toprule
   & \subtrans  &Latency & \#Params & BLEU \\
    \midrule
    \multirow{2}{*}{\shortstack[c]{WMT'14\\En-De}} & Largest & 10.1s &  71M & 28.1  \\
    & Searched \name & \textbf{6.9s} & \textbf{48M}  & \textbf{28.4} \\
    \midrule
     \multirow{2}{*}{\shortstack[c]{WMT'14\\En-Fr}} & Largest & 10.1s &  71M & 41.4 \\
    & Searched \name & \textbf{9.1s} & \textbf{57M} & \textbf{41.8} \\
    
    \bottomrule
\end{tabular}
\tabcaption{The searched~\name compared with the largest \subtrans in the design space. Larger models do not necessarily have better performance. HAT models have lower latency, smaller size, and higher \bleu.}
\vspace{-10pt}
\label{tab:largest}
\end{table}

\name achieves higher \bleu with 1.5\x lower latency and 1.5\x smaller size compared with the largest \subtrans (Table~\ref{tab:largest}). This suggests that larger models do not always provide better performance, and demonstrates the effectiveness of \name.
We also compare the evolutionary search with random search (Figure~\ref{fig:evo_vs_random}). Evolutionary search can find models with lower losses than random search.

\myparagraph{\subtrans Performance Proxy.}
All SubTransformers inside the \supertrans are \emph{uniformly sampled} and thus \emph{equally trained}, so the performance order is well-preserved during training. We conduct experiments to show the effectiveness of the \subtrans performance proxy as in Table~\ref{tab:proxy} and Appendix Figure~\ref{fig:proxy}. The BLEUs of \subtranss with inherited weights and weights trained from-scratch are very close. More importantly, they also have the \emph{same relative performance order}. Therefore, we can rely on the proxy to search high-performance model architecture, significantly reducing the search cost.

\begin{table}[t]

\renewcommand*{\arraystretch}{1.15}
\setlength{\tabcolsep}{3pt}
\small\centering
\footnotesize{
\begin{tabular}{ccc|ccc}
\toprule
\multicolumn{3}{c|}{WMT'14 En-De} & \multicolumn{3}{c}{WMT'14 En-Fr} \\
\multirow{3}{*}{\shortstack[c]{Inherited\\Val Loss}} & \multirow{3}{*}{\shortstack[c]{Inherited\\\bleu}} &  \multirow{3}{*}{\shortstack[c]{From-\\Scratch\\\bleu}} & \multirow{3}{*}{\shortstack[c]{Inherited\\Val Loss}}  & \multirow{3}{*}{\shortstack[c]{Inherited\\\bleu}} &  \multirow{3}{*}{\shortstack[c]{From-\\Scratch\\\bleu}} \\
 & &\\
 & &\\
\midrule
4.71 & 24.9 & 25.8 & 3.92 & 37.4 & 38.8 \\
4.40 & 25.8 & 27.6 & 3.71 & 38.0 & 40.0 \\
4.07 & 26.3 & 28.1 & 3.48 & 39.5 & 41.1 \\
4.02 & 26.7 & 28.2 & 3.46 & 39.6 & 41.4 \\
4.01 & 26.9 & 28.4 & 3.45 & 39.7 & 41.7 \\
\bottomrule
\end{tabular}
}
\tabcaption{The performance of \subtranss with inherited weights are close to those trained from-scratch, and have the same relative performance order.}

\vspace{-10pt}

\label{tab:proxy}

\end{table}

\begin{figure}[t]
    \centering
    \includegraphics[width=\columnwidth]{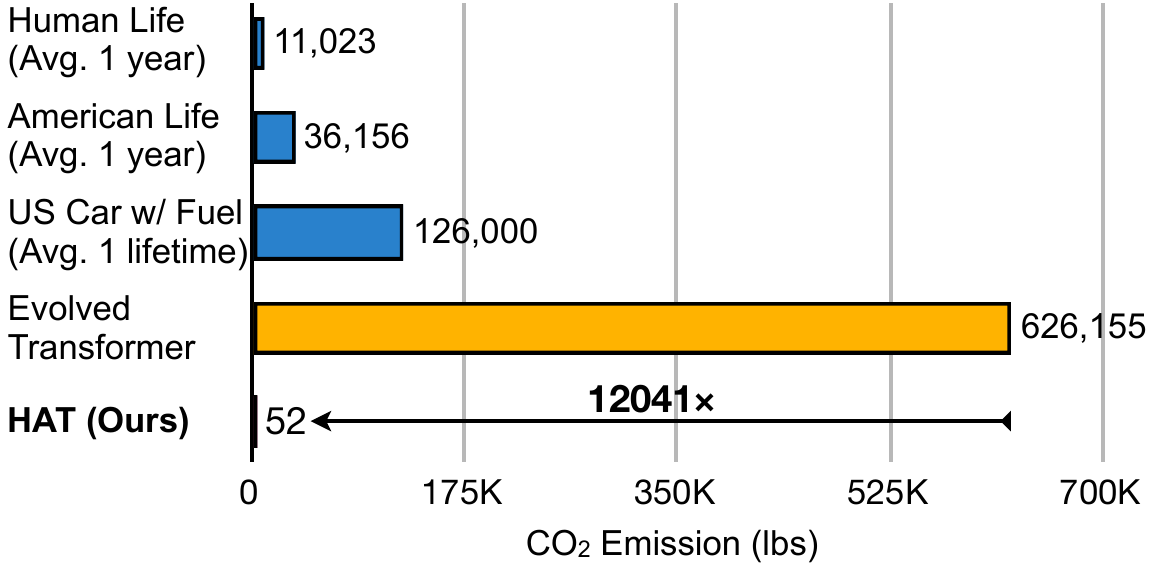}
    \mycaption{The search cost measured in pounds of CO\textsubscript{2} emission. Our framework for searching \name reduces the cost by four orders of magnitude than the Evolved Transformer~\cite{So:2019et}.}
    \label{fig:co2_emission}
\end{figure}

\myparagraph{Low Search Cost.}
As shown in Table~\ref{tab:evo_trans} and Figure~\ref{fig:co2_emission}, the search cost of \name is \lesssearchcostet\x lower than the Evolved Transformer. Although both are using Evolutionary Search, the key difference is that Evolved Transformer needs to train all \emph{individual models} and sort their \emph{final performance} to pick top ones; on the contrary, \name trains \emph{all models together} inside \supertrans and sorts their \emph{performance proxy} to pick top ones. The superior performance of \name proves that the performance proxy is accurate enough to find good models.

\myparagraph{Finetuning Inherited \subtranss} In section~\ref{sec:comp}, we trained each searched \subtrans from-scratch in order to conduct fair comparisons with baselines. In practice, we can also directly finetune the \subtranss with the inherited weights from the \supertrans to further reduce the training cost. With 10K finetuning steps (1/4 of from-scratch training), the inherited \subtranss can achieve similar or better performance than trained from-scratch ones (Table~\ref{tab:finetune}). In this way, the training cost for a model under a new hardware constraint can be further reduced by 4\x, since the \supertrans training cost is amortizable among all searched models.

\begin{table}[t]
\renewcommand*{\arraystretch}{1.15}
\setlength{\tabcolsep}{3pt}
\small\centering
\begin{tabular}{cccc}
    \toprule
 Task   &  From-Scratch 40K  & Inherit-Finetune 10K   \\
    \midrule

\multirow{2}{*}{\shortstack[c]{WMT'14\\En-Fr}} &  41.5 & 41.7 \\
 & 40.0 & 40.2 \\
\midrule

\multirow{2}{*}{\shortstack[c]{WMT'14\\En-De}}  & 28.0 & 28.0 \\
 & 27.5 & 27.4 \\

\bottomrule
\end{tabular}
\tabcaption{The SubTransformer inherited from the SuperTransformer can achieve similar or better performance than the same SubTransformer trained from-scratch. Training steps are saved by 4$\times$.}

\vspace{-10pt}

\label{tab:finetune}
\end{table}

\myparagraph{Quantization Friendly.}
\name is orthogonal to other model compression techniques such as quantization. We apply K-means quantization to \name and further reduce the model size. 
We initialize centroids uniformly in the range of [min, max] of each weight matrix and run at most 300 iterations for each of them. 
Even without any finetuning, 4-bit quantization can reduce the model size by 25\x with negligible \bleu loss compared to the Transformer-Big baseline (\tabref{tab:quant}). Interestingly, the 8-bit model even has 0.1 higher \bleu than the full precision model, indicating the robustness of searched~\name. Compared with the Transformer-Base 4-bit quantization baseline, which has 24MB model size and 38.9 BLEU score, \name has 2.2 higher BLEU with similar model size.

\myparagraph{Knowledge Distillation Friendly.}
\name is also orthogonal to knowledge distillation (KD) because \name focuses on searching for an efficient architecture while KD focuses on better training a given architecture. We combine KD with \name by distilling token-level knowledge (top-5 soft labels) from a high-performance SubTransformer to a low-performance SubTransformer on WMT'14 En-De task. The teacher model has a BLEU of 28.5 and 49M parameters; the student model has 30M parameters. KD can improve the BLEU of the student model from 25.8 to 26.1.

\section{Related Work}
\begin{table}[t]
\renewcommand*{\arraystretch}{1.15}
\setlength{\tabcolsep}{3pt}
\small\centering
\begin{tabular}{lccccc}
    \toprule
   &  BLEU  & Model Size & Reduction   \\
    \midrule
    Transformer Float32 & 41.2& 705MB & -- \\
    \name Float32 & 41.8 &227MB  & 3\x  \\
    \textbf{\name 8 bits} & \textbf{41.9}  & \textbf{57MB} & \textbf{12\x}\\
    \textbf{\name 4 bits} & 41.1  & \textbf{28MB}   & \textbf{25\x}\\
    \bottomrule
\end{tabular}
\tabcaption{K-means quantization of \name models on WMT'14 En-Fr. 4-bit quantization reduces the model size by 25\x with only 0.1 \bleu loss compared with the transformer baseline. 8-bit quantization even has 0.1 higher \bleu than its full precision version.}
\vspace{-17pt}

\label{tab:quant}
\end{table}

\myparagraph{Transformer.}
Transformer~\cite{Vaswani:2017attention} has prevailed in sequence modeling~\cite{ng-etal-2019-facebook, junczys2018microsoft}. By stacking identical blocks, the model obtains a large capacity but incurs high latency. 
Recently, a research trend is to modify the Transformer to improve the performance~\cite{Chen:2018vf, Wu:2019payless, Sukhbaatar:2019adaptive, wang-etal-2019-learning-deep}. 
Among them, ~\citet{Wu:2019payless} introduced a convolution-based module to replace the attention;~\citet{wang-etal-2019-learning-deep} proposed to train deep Transformers by propagating multiple layers together in the encoder. ~\citet{zhang2018accelerating} and~\citet{kim-etal-2019-research} also proposed AAN and SSRU to replace the attention mechanism. \name is orthogonal to them and can be combined to search for efficient architecture with those new modules.
Another trend is to apply non- or partially-autoregressive models to cut down the iteration number for decoding~\cite{Gu:2019levenshtein, akoury-etal-2019-syntactically, wei-etal-2019-imitation,  gu2018nonautoregressive}. Although reducing latency, they sometimes suffer from low performance. \citet{bapna-etal-2018-training} explored using learned linear combinations of encoder outputs as decoder inputs, while \name concatenates the outputs without linear combinations, thus better preserving the low-level information.
\citet{Anonymous:2020efficient} investigated mobile settings for NLP tasks and proposed a multi-branch Lite Transformer. However, it relied on FLOPs for efficient model design, which is an inaccurate proxy for hardware latency (Figure~\ref{fig:latency_profile}). There are also works~\cite{kim2016sequence, junczys-dowmunt-etal-2018-marian, kim-etal-2019-research, micronet} using Knowledge Distillation (KD) to obtain small student models. Our method is orthogonal to KD and can be combined with it to improve the efficiency further. 
There are also hardware accelerators~\cite{ham20203, zhang2020sparch} for attention and fully-connected layers in the Transformer to achieve efficient processing.

\myparagraph{Neural Architecture Search.}
In the computer vision community, there has been an increasing interest in automating efficient model design with Neural Architecture Search (NAS)~\cite{Zoph:2017uo, Zoph:2018ta,Pham:2018tl, He:2018am}.
Some applied black-box optimization such as evolutionary search~\cite{apq} and reinforcement learning~\cite{Cai:2019ui, He:2018am, learncircuits, learncircuits2, mao2019park}; Some leveraged backpropagation with differentiable architecture search~\cite{liu2018darts}. Some also involved hardware constraints into optimizations such as MNasNet~\cite{Tan:2019vw}, ProxylessNAS~\cite{Cai:2019ui}, FBNet~\cite{Wu:2019tk} and APQ~\cite{apq}.  To reduce the NAS cost, supernet based methods~\cite{Pham:2018tl, pmlr-v80-bender18a, Guo:2019single} apply a proxy for sub-network performance and adopt search algorithms to find good sub-networks.
For NLP tasks, the benefits of the architecture search have not been fully investigated.
Recently, \citet{So:2019et} proposed the Evolved Transformer to search for architectures under model size constraints and surpassed the original Transformer baselines. However, it suffered from very high search costs (\emph{250 GPU years}), making it unaffordable to search specialized models for various hardware and tasks. In addition, hardware latency feedback was not taken into account for better case-by-case specializations. Since different hardware has distinct architecture and features~\cite{cong2018understanding}, feedback from hardware is critical for efficient NLP.

\section{Conclusion}

We propose Hardware-Aware Transformers (\name) framework to solve the challenge of efficient deployments of Transformer models on various hardware platforms. We conduct hardware-aware neural architecture search in an ample design space with an efficient weight-shared \supertrans, consuming four orders of magnitude less cost than the prior Evolved Transformer, and discover high-performance low-latency models.
We hope \name can open up an avenue towards efficient Transformer deployments for real-world applications.

\section*{Acknowledgment}
We thank  NSF Career Award \#1943349, MIT-IBM Watson AI Lab, Semi-conductor Research Corporation (SRC), Intel, and Facebook for supporting this research.

\bibliographystyle{acl_natbib}
\bibliography{main.bib}

\newpage
\appendix
\section{Appendix for ``HAT: \underline{H}ardware-\underline{A}ware \underline{T}ransformers for
Efficient Natural Language Processing"}

\subsection{\subtrans Performance Proxy}
\begin{figure}[t]
    \centering
    \includegraphics[width=0.9\columnwidth]{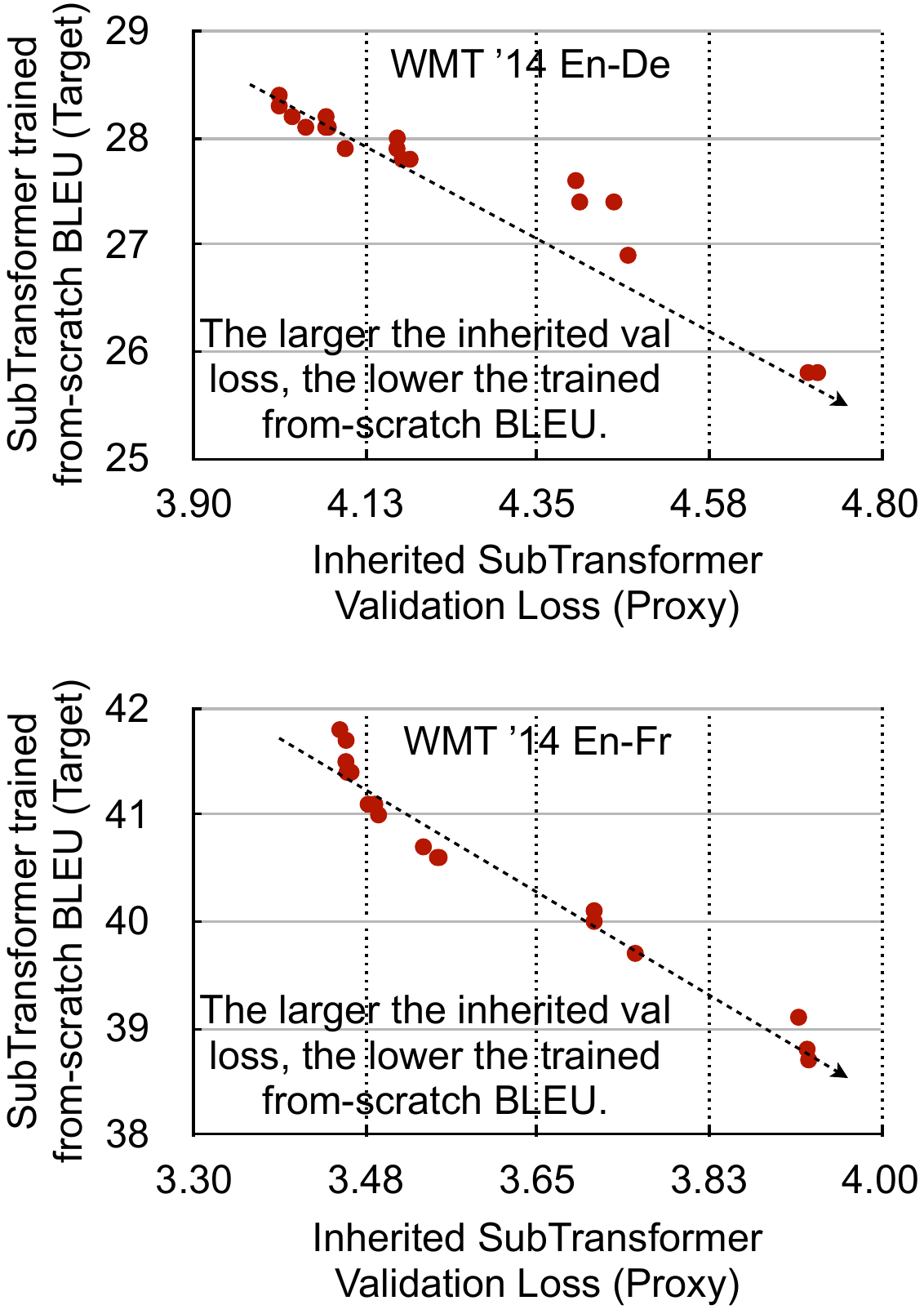}
    \caption{The validation loss of \subtranss is a good performance proxy for \bleu of from-scratch trained \subtranss. The larger the validation loss, the lower the \bleu score.}
    \label{fig:proxy}
\end{figure}

In Figure~\ref{fig:proxy}, we show the relationship between the validation loss of \subtranss directly inherited from the \supertrans, and the \bleu score of the \subtranss trained from-scratch. We can observe that the larger the validation loss, the lower the \bleu score. Therefore the validation loss can be a good performance proxy.

\subsection{Visualizations of Searched Models on WMT'14 En-De Task}
\begin{figure*}[t]
    \centering
    \includegraphics[width=\textwidth]{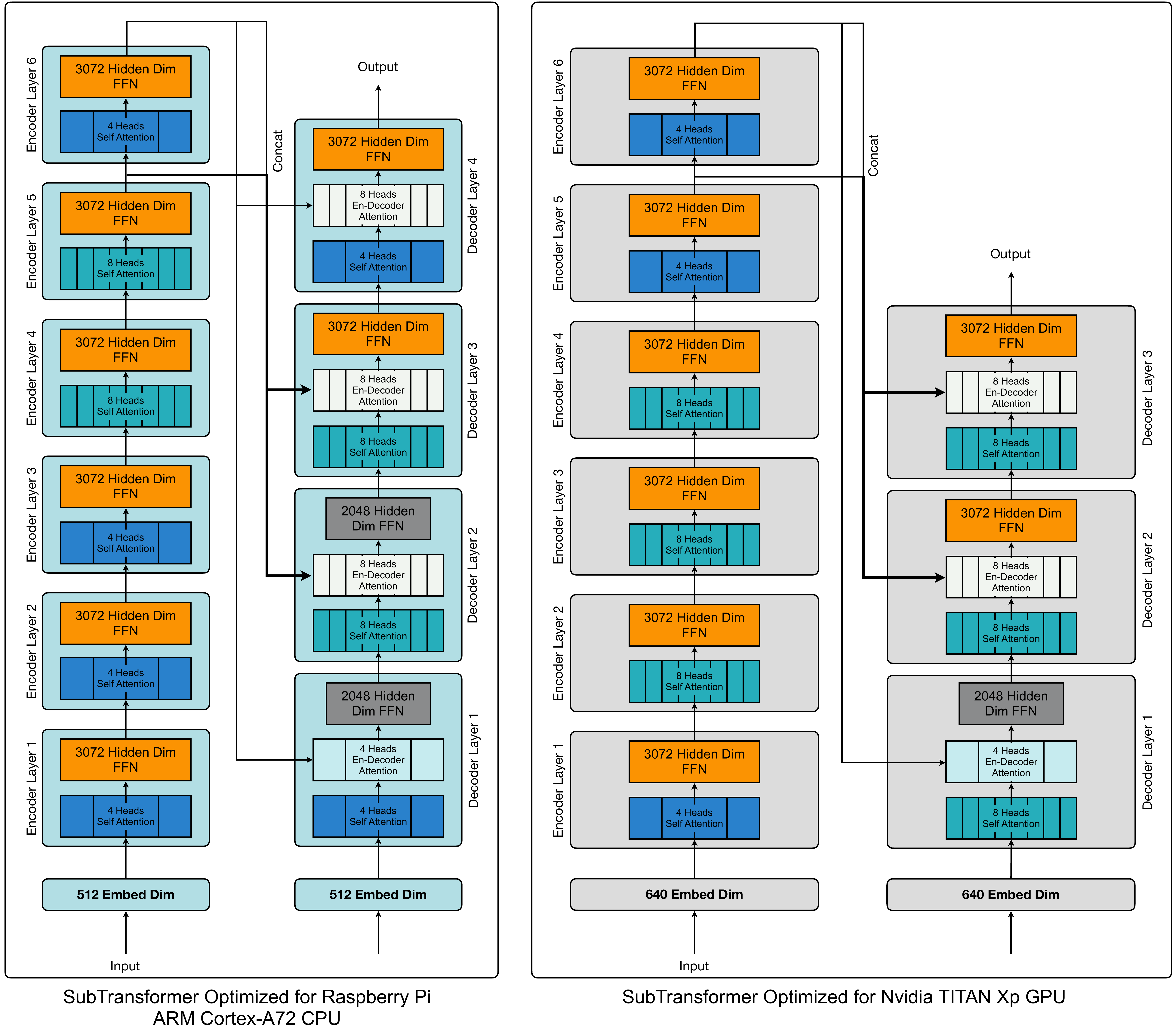}
    \mycaption{\subtranss optimized for Raspberry Pi ARM CPU and Nvidia GPU on WMT'14 En-De task are different. The CPU model has \bleu 28.10, and GPU model has \bleu 28.15.}
    \label{fig:supp}
\end{figure*}

We show the \name models searched for Raspberry Pi ARM Cortex-A72 CPU and Nvidia TITAN Xp GPU in Figure~\ref{fig:supp}. The searched model for Raspberry Pi is deep and thin, while that for GPU is shallow and wide. The \bleu scores of the two models are similar: 28.10 for Raspberry Pi CPU, and 28.15 for Nvidia GPU.

\subsection{Latency, \bleu and \sacre of searched \name models.}
In Table~\ref{tab:appendix}, we show the specific latency numbers, \bleu and \sacre~\cite{post2018call} scores for searched \name models in Figure~\ref{fig:tradeoff_curve} and Figure~\ref{fig:19ende}.

\begin{table}[ht]
\renewcommand*{\arraystretch}{1.15}
\setlength{\tabcolsep}{3pt}
\small\centering
\begin{tabular}{cccccc}
    \toprule
 Task & Hardware  & Latency & \bleu  & \sacre \\
    \midrule

\multirow{17}{*}{\shortstack[c]{\\WMT'14\\En-De}} &  \multirow{6}{*}{\shortstack[c]{Raspberry Pi\\ARM Cortex-A72 \\ CPU}} & 3.5s & 25.8 & 25.6 \\
& & 4.0s & 26.9 & 26.6 \\
& & 4.5s & 27.6 & 27.1 \\
& & 5.0s & 27.8 & 27.2 \\
& & 6.0s & 28.2 & 27.6 \\
& & 6.9s & 28.4 & 27.8 \\
\cmidrule{2-5}
& \multirow{6}{*}{\shortstack[c]{Intel\\ Xeon E5-2640 \\CPU}}   & 137.9ms & 25.8 & 25.6  \\
& & 204.2ms & 27.6 & 27.1 \\
& & 278.7ms & 27.9 & 27.3 \\
& & 340.2ms & 28.1 & 27.5 \\
& & 369.6ms & 28.2 & 27.6 \\
& & 450.9ms & 28.5 & 27.9 \\

\cmidrule{2-5}
& \multirow{5}{*}{\shortstack[c]{Nvidia\\TITAN Xp\\GPU}}   & 57.1ms & 25.8 & 25.6 \\
& & 91.2ms & 27.6 & 27.1 \\
& & 126.0ms & 27.9 & 27.3 \\
& & 146.7ms & 28.1 & 27.5 \\
& & 208.1ms & 28.5 & 27.8 \\

\midrule

\multirow{16}{*}{\shortstack[c]{\\WMT'14\\En-Fr}} &  \multirow{6}{*}{\shortstack[c]{Raspberry Pi\\ARM Cortex-A72\\ CPU}} & 4.3s & 38.8 & 36.0 \\
& & 5.3s & 40.1 & 37.3 \\
& & 5.8s & 40.6 & 37.8 \\
& & 6.9s & 41.1 & 38.3 \\
& & 7.8s & 41.4 & 38.5 \\
& & 9.1s & 41.8 & 38.9 \\

\cmidrule{2-5}
& \multirow{5}{*}{\shortstack[c]{Intel\\ Xeon E5-2640 \\ CPU}}   & 154.7ms & 39.1 & 36.3 \\
& & 208.8ms & 40.0 & 37.2 \\
& & 329.4ms & 41.1 & 38.2 \\
& & 394.5ms & 41.4 & 38.5 \\
& & 442.0ms & 41.7 & 38.8 \\

\cmidrule{2-5}
& \multirow{5}{*}{\shortstack[c]{Nvidia\\TITAN Xp \\ GPU}}  & 69.3ms & 39.1 & 36.3 \\
& & 94.9ms & 40.0 & 37.2 \\
& & 132.9ms & 40.7 & 37.8 \\
& & 168.3ms & 41.1 & 38.3 \\
& & 208.3ms & 41.7 & 38.8 \\

\midrule
\multirow{6}{*}{\shortstack[c]{WMT'19\\En-De}} &  \multirow{6}{*}{\shortstack[c]{Nvidia \\ TITAN Xp\\GPU}}
& 55.7ms & 42.4 & 41.9 \\
& & 93.2ms & 44.4 & 43.9 \\
& & 134.5ms & 45.4 & 44.7 \\
& & 176.1ms & 46.2 & 45.6 \\
& & 204.5ms & 46.5 & 45.7 \\
& & 237.8ms & 46.7 & 46.0 \\

\midrule
\multirow{5}{*}{\shortstack[c]{IWSLT'14\\De-En}} &  \multirow{5}{*}{\shortstack[c]{Nvidia\\ TITAN Xp \\ GPU}}  & 45.6ms & 33.4 & 32.5 \\
& & 74.5ms & 34.2 & 33.3 \\
& & 109.0ms & 34.5 & 33.6 \\
& & 137.8ms & 34.7 & 33.8 \\
& & 168.8ms & 34.8 & 33.9 \\

\bottomrule
\end{tabular}
\tabcaption{Specific latency numbers, \bleu and \sacre scores for searched \name models in Figure~\ref{fig:tradeoff_curve} and Figure~\ref{fig:19ende}.}

\label{tab:appendix}
\end{table}

\end{document}